\documentclass[conference]{IEEEtran}
\IEEEoverridecommandlockouts
\usepackage{cite}
\usepackage{amsmath,amssymb,amsfonts}
\usepackage{algorithmic}
\usepackage{graphicx}
\usepackage{textcomp}
\usepackage{xcolor}
\def\BibTeX{{\rm B\kern-.05em{\sc i\kern-.025em b}\kern-.08em
    T\kern-.1667em\lower.7ex\hbox{E}\kern-.125emX}}

\usepackage{xspace}
\usepackage{subcaption}
\usepackage{colortbl}
\usepackage{float}
\usepackage{booktabs}

\usepackage{wrapfig}
\usepackage{multirow}
\usepackage{enumitem}
\usepackage[hidelinks]{hyperref}

\def\ie{\emph{i.e.}}
\def\eg{\emph{e.g.}}

\def\vs{\emph{vs.}}

\def\Our{{Tail-SD}\xspace}

\newcommand{\myparagraph}[1]{{\setlength{\parskip}{0.3em} \noindent \textbf {#1}}}

\newcommand{\xinyu}[1]{{\color{black}#1}}

\begin{document}

\title{Improving Long-Tail Detection via Layout–Instruction Conditioned Diffusion Synthesis}

\author{\IEEEauthorblockN{Xinyu Zhang$^1$, Xiaodi Wang$^2$, Yanpeng Sun$^3$, Lingfeng Yang$^4$, Jingdong Wang$^5$, Lingqiao Liu$^6$}
\IEEEauthorblockA{
$^1$ University of Auckland, Auckland, New Zealand\quad $^2$ Dcar, Beijing, China \\
$^3$ Singapore University of Technology and Design, Singapore\quad $^4$ WIZYELAB PTE. LTD., Singapore\\
$^5$ Baidu Inc., Beijing, China \quad $^6$ Adelaide University, Adelaide, Australia \\
xinyu.zhang@auckland.ac.nz; 
wangxiaodigm@gmail.com;
yanpengsun115@gmail.com; ylingfeng98@gmail.com}
}

\maketitle

\begin{abstract}

Synthetic data from diffusion models can improve object detection, but long-tail performance is still limited by scarce rare classes and inflexible conditioning (\eg, text-only or layout-only inputs). In this paper, we propose \Our, a layout–instruction conditioned stable diffusion framework that supports text, layouts, or their hybrid, enabling controllable synthesis from boxes, masks, and/or prompts to better cover the long-tail objects. To train a single model across heterogeneous conditions, we introduce a multi-task cross-attention module and multi-task embedding tokens, improving generation quality and diversity. Extensive experiments on COCO and LVIS demonstrate that diffusion synthesis with hybrid conditioning consistently benefits detection, \eg, on LVIS val, it achieves +4.89\% / +5.68\% rare Box/Mask AP, highlighting strong potential for addressing long-tail challenges.

\end{abstract}


\section{Introduction}
\label{sec:intro}

Stable diffusion models~\cite{ho2020denoising, song2020denoising,rombach2022high,saharia2022photorealistic,ramesh2021zero,ramesh2022hierarchical,betker2023improving,podell2023sdxl,koh2024generating,xie2025dymo,li2026let,xie2026hyperalign,rong2026distillation} have demonstrated significant progress in generative technology, presenting advanced capabilities to generate high-fidelity and realistic images. 
Some works~\cite{hertz2022prompt,tumanyan2023plug,nichol2021glide,avrahami2022blended,couairon2022diffedit,brooks2023instructpix2pix,li2023gligen,fu2023guiding,yeh2024gen4gen,zhang2024magicbrush,sheynin2023emu,yang2024add,you2025images} leverages diffusion-generated images as synthetic data to improve downstream perception tasks~\cite{li2023diffusion,ji2024latent,azizi2023synthetic,he2022synthetic,sariyildiz2023fake,tian2024stablerep,zhao2023x,suri2023gen2det,wasserman2024paint}, including object detection.

Despite these advances, improving \emph{long-tail} object detection remains a challenge.
Real-world datasets are highly imbalanced: rare categories have limited instances and weak coverage of intra-class variations, causing detectors to underfit rare concepts and overfit head classes.
A natural solution is to synthesize additional training samples, but \emph{long-tail} gains require more than ``more images'':
effective augmentation must be \emph{targeted} (to deliberately increase rare-class exposure), \emph{controllable} (to ensure the desired category appears with correct attributes), and \emph{diverse} (to enrich backgrounds, poses, viewpoints, and co-occurrences).
However, existing diffusion-based synthesis pipelines often constrain conditioning to a single fixed format, limiting their ability to generate reliable and diverse rare-category samples.

\begin{figure}
    \centering
    \includegraphics[width=0.95\linewidth]{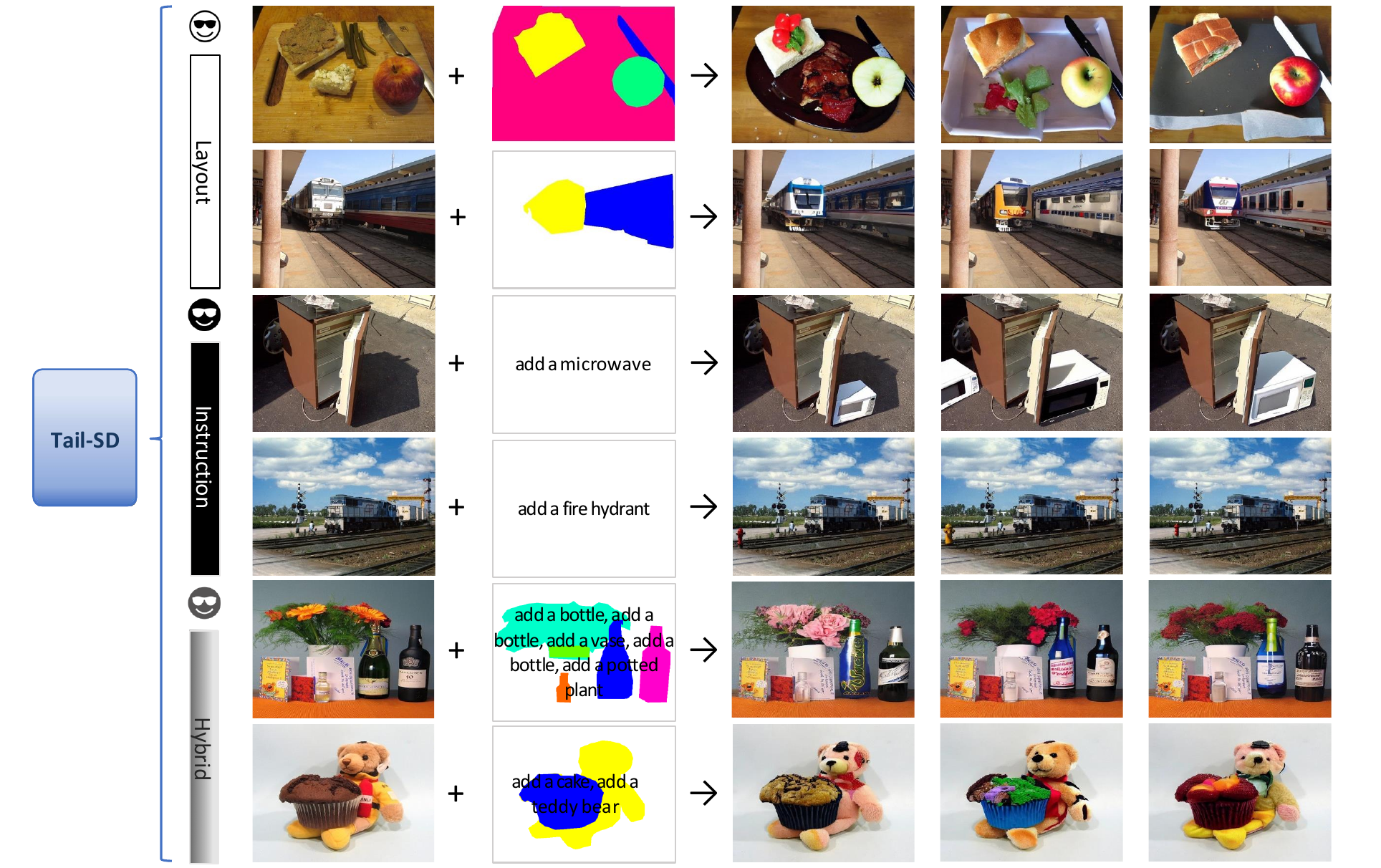}
    \captionof{figure}{Our \Our is a flexible and versatile image generation pipeline, supporting text, layouts and a hybrid of both as inputs. The last three synthetic images are sampled from different seeds. (Zoom in for best view)}
    \label{fig:figure1}
    \vspace{-0.5cm}
\end{figure}

Existing diffusion synthesis methods for detection~\cite{yang2023paint,li2023gligen,chen2023anydoor,zhao2023x,suri2023gen2det,chai2023layoutdm,yeh2024gen4gen,wasserman2024paint,yang2024add} can be broadly categorized into two lines by their input formats: \emph{layout-based} and \emph{instruction-based}.
Layout-based approaches~\cite{yang2023paint,li2023gligen,chen2023anydoor,zhao2023x,suri2023gen2det,xue2023freestyle} take structured layouts (\eg, masks or bounding boxes) to guide object composition and spatial arrangement.
While precise, they often require costly manual layout specification and are hard to scale to many rare categories. Moreover, fixed or simplistic layouts can bottleneck diversity, which is crucial for long-tail generalization.
Recent attempts~\cite{tang2023layoutnuwa,chai2023layoutdm,yeh2024gen4gen,chen2024towards} try to automatically generate layouts, yet maintaining reliability remains difficult, especially in object size, shape, position, and inter-object relationships.
In contrast, instruction-based methods~\cite{fu2023guiding,wasserman2024paint,brooks2023instructpix2pix,zhang2024magicbrush} enable flexible prompting without explicit layouts, but may provide weaker spatial control and inconsistent instruction execution, making it difficult to reliably synthesize rare objects with desired attributes and minimal artifacts.
These limitations motivate a unified framework that can exploit \emph{both} the structural controllability of layouts and the semantic flexibility of instructions.

To this end, we propose \Our, a layout-instruction conditioned stable diffusion framework tailored for long-tail detection.
\Our supports text, layouts, or their hybrid as inputs, allowing users to generate images based on what they \emph{have} (boxes/masks) and what they \emph{want} (instructions), which is particularly useful for targeted rare-class augmentation.
Importantly, \Our trains a single diffusion model that can flexibly adapt to different condition types via multi-task learning, so that robust condition-to-image alignment learned from abundant head classes can regularize and transfer to rare classes, improving the reliability of long-tail targeted synthesis.
We design a \textit{multi-task cross-attention module} together with \textit{multi-task embedding tokens} to unify heterogeneous conditions in a simple yet effective way, improving generation fidelity while enhancing diversity across modalities.

We evaluate \Our on object detection task.
Following standard settings~\cite{ghiasi2021simple,zhao2023x,suri2023gen2det}, we augment detector training data by mixing synthetic images generated by \Our with the original data.
Experiments show consistent improvements on both COCO and LVIS, and most importantly, substantial gains on LVIS rare categories, demonstrating the effectiveness of layout-instruction hybrid diffusion synthesis for addressing long-tail challenges.
Our contributions are summarized as:
\begin{itemize}[left=0em]
\item[$\bullet$] We introduce \Our, a layout-instruction conditioned stable diffusion framework that supports text, layouts, or hybrid inputs for targeted and diverse synthetic augmentation in long-tail detection.
\item[$\bullet$] We propose a multi-task cross-attention module and multi-task embedding tokens to enable effective multi-task learning across heterogeneous conditioning modalities within a single diffusion model.
\item[$\bullet$] Extensive experiments on COCO and LVIS demonstrate consistent detection gains, with particularly strong improvements on rare categories, highlighting the potential of \Our for long-tail challenges.
\end{itemize}

\section{\Our}
As in Introduction, for long-tail augmentation, we often need targeted, controllable and diverse rare-class samples. This raises a key question: \textit{can we build one diffusion model that supports layout conditioning, instruction conditioning, and their hybrid, and jointly learns from them to better serve rare categories?}
We answer this with \Our, a layout-instruction conditioned stable diffusion framework trained with multi-task learning.
The core idea is to couple complementary signals (layout for spatial grounding, and instruction for semantic guidance) within a single model so that robust condition-to-visual alignment learned from abundant head-class supervision can \emph{regularize} and \emph{transfer} to rare classes, improving the reliability of long-tail targeted synthesis.

\begin{figure*}[t]
    \centering
    \setlength{\fboxrule}{0pt}

    \begin{subfigure}[t]{0.62\linewidth}
        \centering
        \fbox{\includegraphics[width=1.0\linewidth]{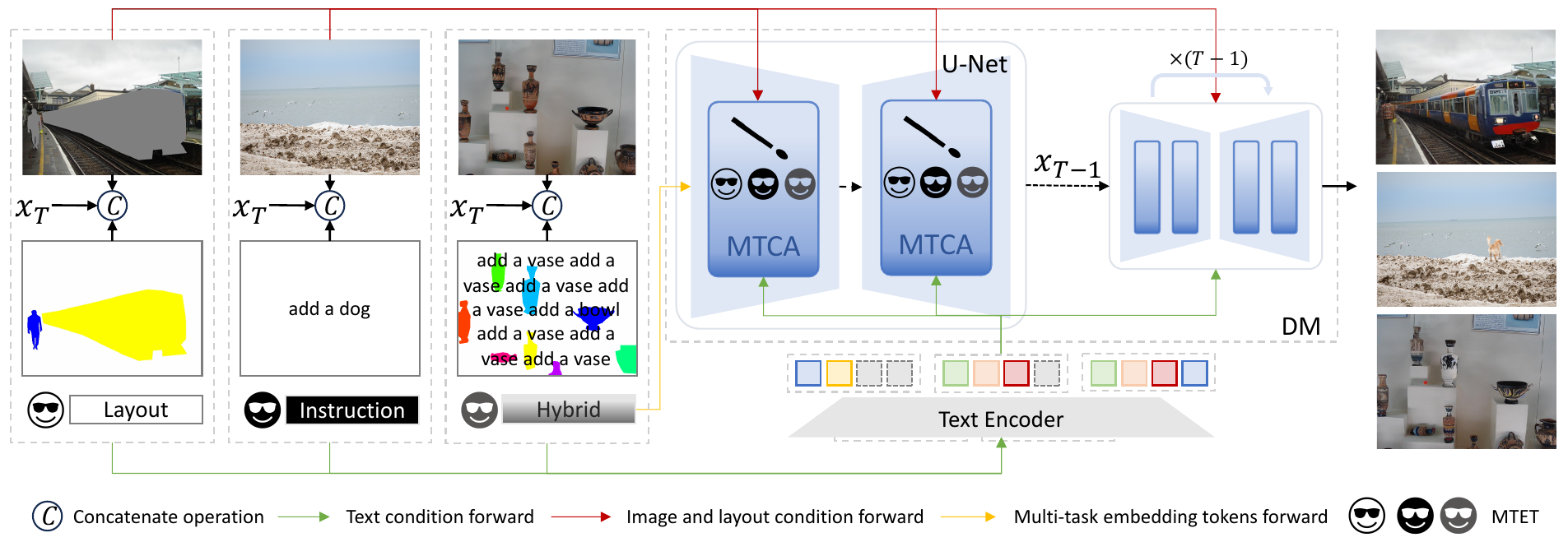}}
        \vspace{-0.4cm}
        \caption{}
        \label{fig:tailsd}
    \end{subfigure}\hfill
    \begin{subfigure}[t]{0.36\linewidth}
        \centering
        \includegraphics[width=1.0\linewidth]{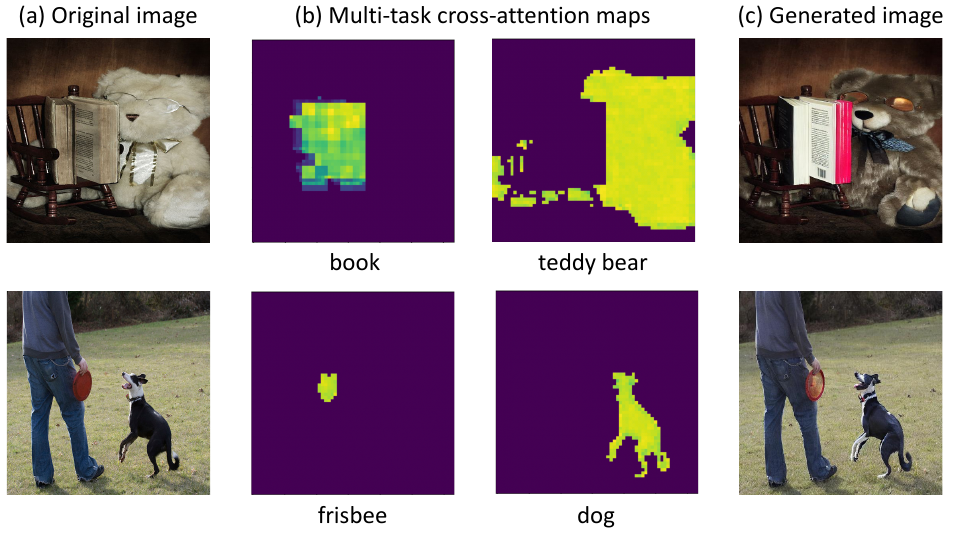}
        \vspace{-0.35cm}
        \caption{}
        \label{fig:attn_map}
    \end{subfigure}

    \vspace{-0.25cm}
    \caption{
    {(a)} Overview of \Our. The model supports instruction, layout, and hybrid conditioning, including two components: (i) MTCA that applies format-aware attention control, and (ii) MTET that indicates the conditioning mode to the diffusion model. The autoencoder is omitted for clarity.
    {(b)} Visualization of multi-task cross-attention maps in our \Our: for each original image, we show two MTCA maps emphasizing corresponding objects and the generated image.
    }
    \label{fig:tailsd_attn}
    \vspace{-0.5cm}
\end{figure*}

\subsection{Framework}
The overview pipeline of \Our is shown in Fig.~\ref{fig:tailsd}.
\Our extends a standard latent diffusion model with two modules, \ie, multi-task cross attention module (MTCA) (Sec.~\ref{sec:multi-task cross attention}) and multi-task embedding tokens (MTET) (Sec.~\ref{sec:multi-task embedding token}) to enable unified training and inference across three conditioning modes:
\emph{Instruction} (text-only),  \emph{Layout} (boxes/masks with class names), and \emph{Hybrid} (layout + instruction).
This design directly matches long-tail augmentation needs: users can synthesize images from what they \emph{have} (boxes/masks for rare categories) and what they \emph{want} (semantic instructions for diversity).

\subsection{Multi-task cross attention module (MTCA)}\label{sec:multi-task cross attention}
In original stable diffusion, cross-attention takes noisy latent features as queries and text embeddings as keys/values:
$\mathrm{CrossAttn}(Q, K, V) = \mathcal{A} \cdot V = \mathrm{softmax}(\frac{QK^T}{\sqrt{d}})\cdot V$, 
where $\mathcal{A}$ is the attention map and $d$ is the key/query dimension.
Different channels in cross-attention maps can somewhat reflect the specific text prompt~\cite{hudson2021generative,hertz2022prompt,liu2024towards}. Building on this property, \cite{xue2023freestyle} enforces spatial control by rectifying the attention map of each token with its associated mask, thereby grounding token attention to user-specified regions.
Inspired by this, we present a multi-task cross attention module (MTCA) that generalizes attention rectification to multiple conditioning modes: when a layout is provided, MTCA spatially gates token attention to satisfy the layout constraint; when no layout is available, it degenerates to standard cross-attention so the model can fully follow the instruction prompt.

Let $\mathcal{L}\in\mathbb{R}^{C\times H\times W}$ denote a layout given by a user, where each channel $\mathcal{L}^k$ is a binary map indicating the region of the $k$-th object (from a mask or a box).
We align layout channels to a subset of text tokens that correspond to object concepts (\eg, category words); tokens without layout supervision (\eg, padding or non-object words) use an all-one map $J\in\mathbb{R}^{H\times W}$.
We introduce a simple gate $\sigma\in\{0,1\}$ indicating whether layout is provided:
$\sigma=1$ for \emph{layout} and \emph{hybrid} modes; $\sigma=0$ for \emph{instruction} mode.
We rewrite the effective layout as
\begin{equation}
\mathcal{L}^k \leftarrow \sigma\cdot \mathcal{L}^k + (1-\sigma)\cdot J.
\label{eq:layout_gate}
\vspace{-0.05cm}
\end{equation}
When no layout is available, $\mathcal{L}$ becomes all-ones and MTCA reduces to standard cross-attention.

We apply $\mathcal{L}$ to rectify cross-attention maps, enforcing spatial focus for object-related tokens:
\begin{equation}
\widehat{\mathcal{A}}=\mathcal{A}\odot \mathcal{L},\quad
\widehat{\mathcal{A}}^k_{i,j}=
\begin{cases}
\mathcal{A}^k_{i,j}, & \mathcal{L}^k_{i,j}=1,\\
0, & \mathcal{L}^k_{i,j}=0,
\end{cases}
\label{eq:mtca}
\vspace{-0.05cm}
\end{equation}
where $\odot$ denotes element-wise multiplication.
$\mathcal{A}^k_{i,j}$ and $\mathcal{L}^k_{i,j}$ denote the value at position $(i,j)$ of the $k$-th channel of $\mathcal{A}$ and $\mathcal{L}$, respectively.
MTCA is thus
$\mathrm{MTCA}(Q,K,V,{\mathcal{L}})=\widehat{\mathcal{A}}\cdot V$.
Intuitively, MTCA provides \emph{precise spatial grounding} when layouts exist (critical for ensuring rare objects appear), while remaining compatible with instruction-only synthesis.
Fig.~\ref{fig:attn_map} illustrates examples of multi-task cross-attention maps.

\subsection{Multi-task embedding tokens}\label{sec:multi-task embedding token}
Although MTCA unifies the computation graph, training across heterogeneous conditioning modes can be unstable: the model may confuse ``no-layout'' samples with ``weak-layout'' cases, and instruction-only supervision may influence or conflict with layout grounding.
To disambiguate modes and stabilize optimization, we introduce learnable multi-task embedding tokens (MTET).
Following~\cite{emu}, we learn a small set of mode-specific embedding tokens $\{v_{\sigma}\}$ that serve as input gates.
In our setting, we use \emph{two} tokens with $\sigma\in\{0,1\}$:
$v_{0}$ corresponds to \emph{layout-driven} conditioning (including {layout-only} and {layout-instruction hybrid}),
and $v_{1}$ corresponds to \emph{instruction-only} conditioning.\footnote{When using layout inputs, we adopt class names as text embeddings as in~\cite{xue2023freestyle}. This makes the layout-only case share the same conditioning structure as the hybrid case, and thus we use the same token $v_0$ for both.} 

Given a training sample $y=(c_T,c_I,x,\sigma)$ consisting of a text prompt $c_T$, an input image $c_I$, a target image $x$, and a mode index $\sigma$, we inject the selected token $v_{\sigma}$ into the diffusion model in two places:
(i) we add $v_{\sigma}$ to the timestep embedding, and
(ii) we concatenate $v_{\sigma}$ to the text embeddings fed into cross-attention in the U-Net.
This provides an explicit mode indicator, improves separation between conditioning formats, and requires no architectural change to the backbone. The training objective is thus:
\begin{equation}
\min_{\theta,\sigma\in\{0,1\}} 
\mathbb{E}_{y,\epsilon,t}
\left[
\left\| \epsilon - \epsilon_{\theta}\!\left(z_t,t, \mathcal{E}(c_I), c_T, v_{\sigma}\right) \right\|_2^2
\right],
\label{eq:loss_task_emb}
\vspace{-0.1cm}
\end{equation}
where $\epsilon\sim\mathcal{N}(0,1)$ is the noise, $\mathcal{E}(\cdot)$ denotes the image encoder, $\theta$ denotes the U-Net parameters, and $z_t$ is the noisy latent of $z=\mathcal{E}(x)$ at timestep $t\in T$.

At inference time, we directly select $v_{\sigma}$ according to the chosen input format (layout/hybrid \vs ~instruction), without any additional mode prediction.
This differs from~\cite{emu}, which predicts task indices using an extra fine-tuned large language model; our approach is simpler and faster, and avoids potential task-misclassification errors.

\myparagraph{Discussion.} 
The works most closely related to ours include layout-driven methods FreestyleNet~\cite{xue2023freestyle} and Gen2Det~\cite{suri2023gen2det}, and instruction-based editing methods such as Emu-Edit~\cite{emu}.
Compared with \cite{xue2023freestyle,suri2023gen2det}, our MTCA is inspired by layout rectification but differs by \emph{unifying} instruction, layout, and hybrid conditioning in a single model, which is crucial for scalable long-tail augmentation.
Compared to \cite{emu}, \Our provides stronger spatial grounding through layouts, improving the reliability of targeted rare-category synthesis.
Finally, while task tokens appear in prior multi-task editing systems~\cite{emu}, our MTET specifically resolves ambiguity across {conditioning formats} (instruction/layout/hybrid) for detection-oriented synthesis, enabling simpler and more stable inference by directly selecting the mode token.

\section{Experiments}
\subsection{Experimental Settings}\label{sec:exp_settings}

\myparagraph{Datasets.} For the training of our \Our, we utilize COCO~\cite{lin2014microsoft}, LVIS~\cite{gupta2019lvis}, Visual Genome~\cite{krishna2017visual}, and VGPhraseCut~\cite{vgcut}.
\textit{Data pre-processing.} For instruction-based tasks, we follow the pipeline in \cite{yang2024add} to create ``removal-instruction-original'' triples.
For the layout-based task, we adopt the approach from \cite{xue2023freestyle}, where each semantic category (\eg, an object class) is represented by a single word (\eg, ``cat'') or a phrase (\eg, ``teddy bear'') to describe the concept. Each concept is associated with its own layout, \ie, a binary map, which is utilized in MTCA.
\textit{Data post-processing.} For downstream object detection, we generate synthetic images only from COCO and LVIS to ensure fair comparison. For positioning newly added objects, we follow \cite{yang2024add} to use \cite{liu2023grounding} to determine their locations based on object labels as queries.

\myparagraph{Implementation details.}
We adopt SD-v1.5~\cite{rombach2022high} as the base model and fine-tune \Our for 30 epochs.
We set the diffusion steps to 50 during denoising.
For detection/instance segmentation, we follow X-Paste~\cite{zhao2023x} with a 4$\times$ schedule (90k iterations) using Detectron2~\cite{wu2019detectron2}.
We evaluate using CenterNet2~\cite{zhou2021probabilistic} and Faster R-CNN~\cite{ren2015faster} with a ResNet-50~\cite{he2016deep} backbone.
All experiments are conducted on 8$\times$  V100 GPUs.

\myparagraph{Synthetic data for detection.}
We mix synthetic images with real images during detector training following~\cite{suri2023gen2det}.
Each mini-batch contains either all real or all synthetic images, sampled with a fixed ratio.
Unless otherwise specified, we use a 20\% synthetic ratio (real:synth = 4:1).

\myparagraph{Evaluation metrics.}
1) \textit{Detection.} On COCO val we report Box/Mask AP (AP$_b$/AP$_m$). On LVIS val we report Box AP and Mask AP and the long-tail splits AP$_\mathrm{r}$/AP$_\mathrm{c}$/AP$_\mathrm{f}$ (rare/common/frequent), with AP$_\mathrm{r}$ as the primary long-tail metric.
2) \textit{Image synthesis quality.}
We follow prior image-editing evaluations and use both pixel-level and feature-level metrics.
For pixel-level measurements, we report {L1} and {L2} distances between the generated image and the ground-truth target image (lower is better).
For feature-level measurements, we compute cosine similarity between image features extracted by {CLIP}~\cite{radford2021learning} and {DINO}~\cite{caron2021emerging} for the generated image and the ground-truth target image, denoted as {CLIP-I} and {DINO} (higher is better).
For instruction-based inputs, we additionally report {CLIP-T}, the CLIP text-image similarity between generated image and instruction ``add a $<$obj$>$'' (higher is better). 
We randomly select 2,000 images from COCO for evaluation.

\begin{table}[t]
    \centering
    \resizebox{0.8\linewidth}{!}{
    \renewcommand{\arraystretch}{1.0}
    \setlength{\tabcolsep}{16pt}
    \begin{tabular}{lcc}
        \toprule
        Method & AP$_b$ & AP$_m$ \\
        \midrule
        Baseline &  46.00 & 39.80 \\
        \ \ + Copy-Paste~\cite{ghiasi2021simple} &  46.40 & 39.80 \\
        \ \ + X-Paste~\cite{zhao2023x} &  46.60 & 39.90 \\
        \ \ + FreeStyleNet~\cite{xue2023freestyle} & 46.67 &  40.33 \\
        \ \ + Gen2Det~\cite{suri2023gen2det} & \textbf{47.18} &  40.44 \\
        \ \ + \Our (ours) & \xinyu{47.06} &  \textbf{40.60} \\
        \bottomrule
    \end{tabular}
    }
    \vspace{-0.2cm}
    \caption{Comparison on COCO val using CenterNet2. 
    Synthetic images are used to augment detector training.}
    \label{tab:coco_comparison}
\end{table}

\begin{table}[t!]
    \centering
    \resizebox{\linewidth}{!}{
    \renewcommand{\arraystretch}{1.0}
    \setlength{\tabcolsep}{2pt}
    \begin{tabular}{lcccccccc}
        \toprule
        \multirow{2.5}{*}{Method}  & \multicolumn{4}{c}{Box} & \multicolumn{4}{c}{Mask} \\
        \cmidrule(lr){2-5} \cmidrule(lr){6-9}
         & AP & AP$_r$ & AP$_c$ & AP$_f$ & AP & AP$_r$ & AP$_c$ & AP$_f$ \\
        \midrule
        Baseline & 33.80 & 20.84 & 32.84 & 40.58 & 29.98 & 18.36 & 29.64 & 35.46 \\
        \ \ + InstructPix2Pix~\cite{brooks2023instructpix2pix} & 24.88 & 17.40 & 24.00 & 29.16 & 22.13 & 15.49 & 21.60 & 25.64  \\
        \ \ + Copy-Paste~\cite{ghiasi2021simple} & 34.31 & 21.19 & 34.32 & 40.07 & 30.29 & 19.97 & \textbf{30.43} & 34.67  \\
        \ \ + X-Paste~\cite{zhao2023x} &34.34 & 21.05 & \textbf{33.86} & \textbf{40.71} & 30.18 & 18.77 & 30.11 & 35.28 \\
        \ \ + Gen2Det~\cite{suri2023gen2det} & 34.70 & 23.78 & 33.71 & 40.61 & 30.82 & 21.24 & 30.32 & \textbf{35.59} \\
        \ \ + GLIGEN~\cite{li2023gligen} & 33.84 & 24.93 & 33.15 & 38.54 & 29.99 & 22.11 & 29.83 & 33.64 \\
        \midrule
        \ \ + \Our (ours) & \textbf{34.97} & \textbf{25.73} & 33.69 & 40.47 & \textbf{31.15} & \textbf{24.04} & 30.21 & 35.32 \\
        \bottomrule
    \end{tabular}
    }
    \vspace{-0.2cm}
    \caption{Comparison on LVIS val using CenterNet2. 
    For InstructPix2Pix~\cite{brooks2023instructpix2pix}, we generate synthetic images using the same instruction templates as \Our for a fair comparison. }
    \label{tab:lvis_comparison}
    \vspace{-0.5cm}
\end{table}

\subsection{Main results}

\myparagraph{Comparison on object detection.}
We evaluate \Our with 
prior synthetic methods~\cite{ghiasi2021simple,zhao2023x,suri2023gen2det,xue2023freestyle,li2023gligen},
\footnote{Note that we re-implement X-Paste using the official code as in the open issue (\url{https://github.com/yoctta/XPaste/issues/2}) and the appendix of \cite{suri2023gen2det}.}.
\textit{Baseline} uses real images only.
Following~\cite{suri2023gen2det}, we train detectors with a mixture of real and synthetic images.
Tab.~\ref{tab:coco_comparison} shows that augmenting training with images generated by \Our yields clear improvements on COCO: \Our improves the baseline by \xinyu{+1.06} Box AP and \xinyu{+0.80} Mask AP.
\Our achieves the best Mask AP and a comparable Box AP to the strongest prior method \cite{suri2023gen2det}.
Notably, while Gen2Det primarily relies on layout inputs, \Our additionally supports instruction conditioning, enabling more flexible and practical targeted synthesis, which becomes particularly beneficial in long-tail settings.
We further evaluate \Our on the LVIS benchmark~\cite{gupta2019lvis}, which splits categories into frequent/common/rare based on instance frequency.
As shown in Tab.~\ref{tab:lvis_comparison}, \Our improves overall performance and delivers substantial gains on rare categories, outperforming layout-only baselines~\cite{zhao2023x,suri2023gen2det} while remaining competitive on common and frequent classes.
In particular, \Our improves the baseline by +1.17 Box AP and +4.89 rare Box AP, and surpasses Gen2Det by +1.95 rare Box AP and +2.80 rare Mask AP.
These results demonstrate that \Our produces effective synthetic data for detection, with pronounced benefits on long-tailed categories.

\begin{table}[t]
    \scriptsize
    \centering
    \resizebox{1.0\linewidth}{!}{
    \renewcommand{\arraystretch}{1.0}
    \setlength{\tabcolsep}{3.5pt}
    \begin{tabular}{llccccc}
        \toprule
         & Method & L1 $\downarrow$ & L2 $\downarrow$ & CLIP-I $\uparrow$ & CLIP-T $\uparrow$ & DINO $\uparrow$ \\
        \midrule
        \multirow{4}{*}{\rotatebox{90}{Instruction}} & InstructPix2Pix~\cite{brooks2023instructpix2pix} & 0.302 & 0.311 & 0.863 & 0.227 & 0.780 \\
        & MagicBrush~\cite{zhang2024magicbrush} & 0.238 & 0.259 & 0.900 & 0.225 & 0.841 \\
        & MGIE~\cite{fu2023guiding} &  0.325 &  0.421  & 0.861   &  \textbf{0.228}  &  0.783   \\
        & \Our (ours) & \textbf{0.195} & \textbf{0.161} & \textbf{0.955} & 0.221 & \textbf{0.970} \\
        \midrule
        \multirow{2}{*}{\rotatebox{90}{\parbox{0.61cm}{\centering Layout}}} & FreestyleNet~\cite{xue2023freestyle} & 0.946 & 1.448 & 0.762 & - & 0.633 \\
        & \Our (ours) & \textbf{0.355} & \textbf{0.413} & \textbf{0.891} & - & \textbf{0.865} \\
        \bottomrule
    \end{tabular}}
    \vspace{-0.2cm}
    \caption{Quantitative comparison on image editing with instruction-based and layout-based inputs.
    }
    \label{tab:methods_comparison_metrics}
    \vspace{-0.2cm}
\end{table}

\begin{figure}[t]
    \centering
    \setlength{\fboxrule}{0pt}
    \fbox{\includegraphics[width=0.95\linewidth]{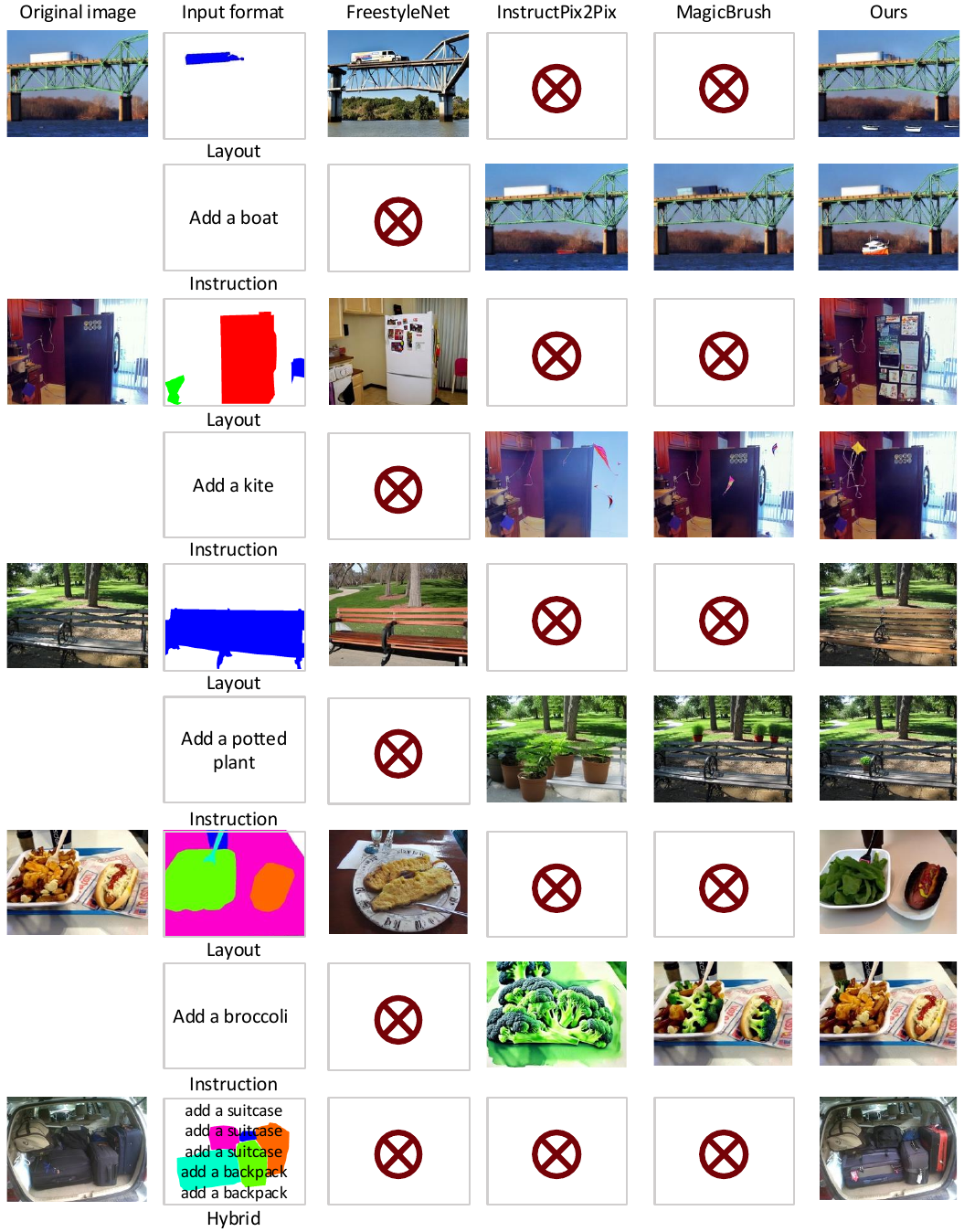}}
    \vspace{-0.3cm}
    \caption{Qualitative comparison of \Our with representative image-editing baselines under different conditioning formats. \Our supports both layout- and instruction-based control while better preserving background consistency, producing more realistic edits. (Best viewed in color and with zoom.)}
    \label{fig:compare_with_others}
    \vspace{-0.5cm}
\end{figure}

\myparagraph{Comparison with image-editing baselines.}
We compare \Our with representative image-editing methods, including instruction-based approaches InstructPix2Pix~\cite{brooks2023instructpix2pix}, MagicBrush~\cite{zhang2024magicbrush}, and MGIE~\cite{fu2023guiding}, as well as the layout-based method FreestyleNet~\cite{xue2023freestyle}.
Tab.~\ref{tab:methods_comparison_metrics} shows that \Our achieves the best overall performance on most metrics under instruction inputs, obtaining the lowest pixel errors (L1/L2) and the highest feature-level similarities (CLIP-I and DINO) to the ground-truth target images.
Although \Our is slightly worse on CLIP-T, this metric mainly measures text-image alignment to the object mention rather than overall fidelity to the full edited target, which can favor methods that over-emphasize the inserted object.
Compared to MGIE~\cite{fu2023guiding}, \Our improves DINO by 0.183, indicating closer perceptual similarity to the ground-truth edited results.
For layout-based inputs, \Our substantially outperforms FreestyleNet~\cite{xue2023freestyle} across all reported metrics.
This is largely because \cite{xue2023freestyle} relies on panoramic segmentation layouts that tend to alter background content (see Fig.~\ref{fig:compare_with_others}), leading to large pixel deviations from the target.
In contrast, \Our better preserves the original background while editing the specified objects, producing more realistic and faithful results.

\begin{table}[t]
    \centering
    \resizebox{0.95\linewidth}{!}{
    \renewcommand{\arraystretch}{1.0}
    \setlength{\tabcolsep}{8pt}
    \begin{tabular}{lccccc}
        \toprule
        \multirow{2.5}{*}{Method} & \multirow{2.5}{*}{Task} & \multicolumn{2}{c}{COCO} & \multicolumn{2}{c}{LVIS} \\
        \cmidrule(r){3-4}
        \cmidrule(r){5-6}
        & & AP$_b$ & AP$_m$ & AP & AP$_r$ \\
        \cmidrule(r){1-6}
        Baseline & - & 46.00 & 39.80 & 33.80 & 20.84 \\
        \multirow{3}{*}{\ \ + \Our} & instruction & 46.80 &  40.36 & 33.96 & 22.26 \\
        & layout & 46.58 &  40.37 & 34.05 & 19.81 \\
        & multi-task & \textbf{47.06} &  \textbf{40.60} & \textbf{34.97} & \textbf{25.73} \\
        \bottomrule
    \end{tabular}
    }
    \vspace{-0.1cm}
    \caption{Effect of multi-task learning on detection  (CenterNet2~\cite{zhou2021probabilistic}). ``instruction'' and ``layout'' denote single-format training, while ``multi-task'' denotes multi-task training.}
    \label{tab:multi-task_learning}
\end{table}

\begin{table}[t]
    \centering
    \resizebox{0.95\linewidth}{!}{
    \renewcommand{\arraystretch}{1.0}
    \setlength{\tabcolsep}{10pt}
    \begin{tabular}{lcccc}
        \toprule
        \multirow{2.5}{*}{Method}  & \multicolumn{2}{c}{COCO} & \multicolumn{2}{c}{LVIS} \\
        \cmidrule(r){2-3}
        \cmidrule(r){4-5}
         & AP$_b$ & AP$_m$ & AP & AP$_r$ \\
        \cmidrule(r){1-5}
        Baseline & 46.00 & 39.80 & 33.80 & 20.84 \\
        \ \ + \Our w/o MTET & 47.02 &  \xinyu{40.43} & \xinyu{33.98} & \xinyu{20.51} \\
        \ \ + \Our  & \textbf{47.06} &  \textbf{40.60} & \textbf{34.97} & \textbf{25.73} \\
        \bottomrule
    \end{tabular}
    }
    \vspace{-0.1cm}
    \caption{Effect of MTET on detection  (CenterNet2~\cite{zhou2021probabilistic}).}
    \label{tab:multi-task token}
    \vspace{-0.5cm}
\end{table}

\begin{table}[t]
    \centering
    \resizebox{0.95\linewidth}{!}{
    \renewcommand{\arraystretch}{1.0}
    \setlength{\tabcolsep}{9pt}
    \begin{tabular}{cccccc}
        \toprule
         \multicolumn{2}{l}{\multirow{1}{*}{Data utilization}} &  \multicolumn{2}{c}{Box}&  \multicolumn{2}{c}{Mask}\\
         \cmidrule[\cmidrulewidth](l){1-2}
         \cmidrule[\cmidrulewidth](l){3-4}
         \cmidrule[\cmidrulewidth](l){5-6}
         instruction & layout & $\text{AP}$ & $\text{AP}_{50}$ & $\text{AP}$ & $\text{AP}_{50}$  \\
         \midrule
         $\times$ & $\times$ & 46.00 & - & 39.80 & - \\
         $\checkmark$ & $\times$ & 46.51 & 65.40 & 40.24 & 62.22 \\
         $\times$ & $\checkmark$ & 46.72 & 65.71 & 40.33 & 62.58 \\
         $\checkmark$ & $\checkmark$ & \textbf{47.06} & \textbf{66.01} & \textbf{40.60} & \textbf{62.66} \\
        \bottomrule
    \end{tabular}
    }
    \vspace{-0.2cm}
    \caption{Effect of synthetic images generated with different conditioning formats on COCO~\cite{lin2014microsoft} using CenterNet2~\cite{zhou2021probabilistic}.}
    \label{tab:data_proportion_instruct_layout}
    \vspace{-0.2cm}
\end{table}

\begin{table}[t!]
    \centering
    \resizebox{0.95\linewidth}{!}{
    \renewcommand{\arraystretch}{1.0}
    \setlength{\tabcolsep}{5pt}
    \begin{tabular}{lccccc}
        \toprule
         \multirow{2}{*}{Model} & \multirow{2}{*}{Method} & \multicolumn{2}{c}{Box}&  \multicolumn{2}{c}{Mask}\\
         \cmidrule[\cmidrulewidth](l){3-4}
         \cmidrule[\cmidrulewidth](l){5-6}
          &  & $\text{AP}$ & $\text{AP}_{50}$ & $\text{AP}$ & $\text{AP}_{50}$  \\
         \midrule 
         \multirow{2}{*}{CenterNet2~\cite{zhou2021probabilistic}} & Baseline & 46.00 & - & 39.80 & - \\ 
        & \ \ + \Our & \textbf{47.06} & \textbf{66.01} & \textbf{40.60} & \textbf{62.66} \\
         \multirow{2}{*}{Faster R-CNN~\cite{ren2015faster}} & Baseline & 40.22 & 61.02 & - & - \\
         & \ \ + \Our  & \textbf{40.47} & \textbf{61.09} & - & - \\
        \bottomrule
    \end{tabular}
    }
    \vspace{-0.1cm}
    \caption{Effect of \Our synthetic augmentation across different detectors on COCO~\cite{lin2014microsoft}.}
    \label{tab:detector}
    \vspace{-0.5cm}
\end{table}

\myparagraph{Qualitative evaluation.}
To qualitatively assess \Our, we compare its editing results with representative baselines: the layout-based FreestyleNet~\cite{xue2023freestyle} and the instruction-based InstructPix2Pix~\cite{brooks2023instructpix2pix} and MagicBrush~\cite{zhang2024magicbrush}.
As shown in Fig.~\ref{fig:compare_with_others}, \Our can follow both layout and instruction inputs, and further supports their hybrid combination.
Compared to instruction-only methods, \Our produces more coherent object insertion with fewer artifacts while maintaining background fidelity,
\eg, \Our generates a more realistic and well-situated potted plant than \cite{brooks2023instructpix2pix,zhang2024magicbrush}. 
When using layout inputs, \Our can preserve background elements (\eg, trees) when modifying the bench, whereas \cite{xue2023freestyle} tends to alter the background due to its panoramic layout representation.
Overall, these visual comparisons highlight that \Our achieves controllable and realistic edits across conditioning formats while better preserving the original image content.

\subsection{Ablation study}

\myparagraph{Effect of multi-task learning.}
To verify the benefit of multi-task learning in \Our, we train single-format variants using only instruction or layout inputs (each trained with its corresponding data). We remove MTET since the input mode is fixed.
As shown in Tab.~\ref{tab:multi-task_learning}, jointly training on both formats consistently improves detection performance, with the largest gain on LVIS rare classes, indicating that multi-task learning produces more effective synthetic data for long-tail detection.

\myparagraph{Effect of MTET.}
We introduce multi-task embedding tokens (MTET) to indicate the conditioning mode and reduce interference between instruction/layout training.
As shown in Tab.~\ref{tab:multi-task token}, removing MTET slightly affects COCO but significantly degrades LVIS performance: MTET improves LVIS by +0.99 AP and boosts rare categories by +5.22 AP$_r$ compared to \Our w/o MTET, demonstrating that MTET is crucial for long-tail gains.
We do not ablate MTCA, since without layout-aware attention rectification the model cannot properly ground different instances, making the comparison ill-defined.

\myparagraph{Effect of generated images with different input formats.}
At inference time, \Our can synthesize two types of images depending on whether layout inputs are used.
We generate each type separately and use them to augment detector training on COCO.
As shown in Tab.~\ref{tab:data_proportion_instruct_layout}, both types of generated images are beneficial, and combining them achieves the best performance across Box/Mask AP and AP$_{50}$, indicating complementary gains from the two conditioning formats.

\myparagraph{Generalization across detectors.}
We evaluate whether synthetic images generated by \Our consistently benefit different detection architectures.
As shown in Tab.~\ref{tab:detector}, augmenting COCO training with our generated images improves both CenterNet2~\cite{zhou2021probabilistic} and Faster R-CNN~\cite{ren2015faster}, demonstrating that the gains are not detector-specific.

\section{Conclusion}~\label{sec:conclusion}
This paper presents \Our, a layout-instruction conditioned diffusion framework designed to improve long-tail object detection via controllable synthetic data generation.
\Our supports instruction-only, layout-conditioned, and hybrid inputs, enabling targeted synthesis for rare categories while maintaining high diversity in appearance and context.
To unify heterogeneous conditions within a single model, we introduce a multi-task cross-attention module and multi-task embedding tokens, which stabilize multi-format training and improve condition-to-image alignment.
Extensive experiments on COCO and LVIS show consistent gains in detection and instance segmentation, with particularly strong improvements on rare categories, demonstrating the effectiveness of hybrid-conditioned diffusion synthesis for long-tail challenges.

\myparagraph{Limitations and societal impact.}
\label{sec:limitation_social_impact}
Our findings are based on public datasets and pretrained diffusion models, which may limit their generalization. Future research could explore more high-quality datasets and various pretrained models to further validate our method. 

{\bibliographystyle{IEEEbib}
\bibliography{icme2026references}
}

\end{document}